\documentclass[conference]{IEEEtran}
\IEEEoverridecommandlockouts
\usepackage{cite}
\usepackage{amsmath,amssymb,amsfonts}
\usepackage{algorithmic}
\usepackage{orcidlink}
\usepackage{graphicx}
\usepackage{textcomp}
\usepackage{xcolor}
\usepackage{booktabs}
\usepackage{subfig}
\usepackage{microtype}
\usepackage{amsmath,amssymb}
\usepackage{float}
\usepackage{pgfplots}
\DeclareUnicodeCharacter{2212}{−}
\usepgfplotslibrary{groupplots,dateplot}
\usetikzlibrary{patterns,shapes.arrows}
\pgfplotsset{compat=newest}

\hypersetup{
    pagebackref=true,
    breaklinks=true,
    colorlinks,
    linkcolor={red!75!black},
    citecolor={blue!75!black},
    urlcolor={blue!75!black},
}
\usepackage[capitalise]{cleveref}

\usepackage{xspace}
\makeatletter
\DeclareRobustCommand\onedot{\futurelet\@let@token\@onedot}
\newcommand{\@onedot}{\ifx\@let@token.\else.\null\fi\xspace}
\makeatother

\newcommand{\ie}{i.\,e.,\xspace}
\newcommand{\eg}{e.\,g.,\xspace}

\def\BibTeX{{\rm B\kern-.05em{\sc i\kern-.025em b}\kern-.08em
    T\kern-.1667em\lower.7ex\hbox{E}\kern-.125emX}}
\begin{document}

\title{Handwriting Extraction and Analysis of Signature Lists in Swiss Popular Initiatives\\
}



\author{
\IEEEauthorblockN{
Marco Peer\textsuperscript{1\textdagger}\orcidlink{0000-0001-6843-0830}, 
Thomas Gorges\textsuperscript{2\textdagger}\orcidlink{0009-0007-0573-0992},
Mathias Seuret\textsuperscript{2}\orcidlink{0000-0001-9153-1031},
Vincent Christlein\textsuperscript{2}\orcidlink{0000-0003-0455-3799},
Andreas Fischer\textsuperscript{1}\orcidlink{0000-0003-0069-3436}
}

\vspace{0.2em}

\begin{minipage}[t]{0.45\textwidth}
\centering
\textsuperscript{1}
\textit{AIBEX Group, University of Fribourg}\\
Fribourg, Switzerland
\end{minipage}
\hfill
\begin{minipage}[t]{0.45\textwidth}
\centering
\textsuperscript{2}
\textit{Pattern Recognition Lab, FAU Erlangen-Nürnberg}\\
Erlangen, Germany
\end{minipage}

\thanks{\textsuperscript{\textdagger}These authors contributed equally.}
}

\maketitle

\begin{abstract}
Popular initiatives and referendums are central to Swiss democracy, yet the validation of handwritten signature lists remains a labor-intensive manual process. This paper investigates the potential of automated document analysis methods, including OCR and AI-based handwriting analysis, to support this task. We propose a pipeline combining template-based line segmentation with text recognition and writer retrieval techniques, evaluated on a dataset of 443 handwritten entries from 418 writers. Results show that OCR struggles with out-of-vocabulary handwriting, with a CER of 29.6\% for first names. In contrast, writer retrieval performs more robustly, reaching an mAP of 50.6\%. Furthermore, our experiments indicate that off-the-shelf OCR systems are not sufficiently reliable for transcription of handwritten signature data, particularly for short, out-of-vocabulary entries such as names or addresses. However, writer retrieval methods can effectively identify visually similar entries across signature lists, making them a suitable tool for supporting the detection of potential duplicate submissions based on handwriting similarity.
\end{abstract}


\begin{figure*}
    
    \centering
    \includegraphics[width=\linewidth]{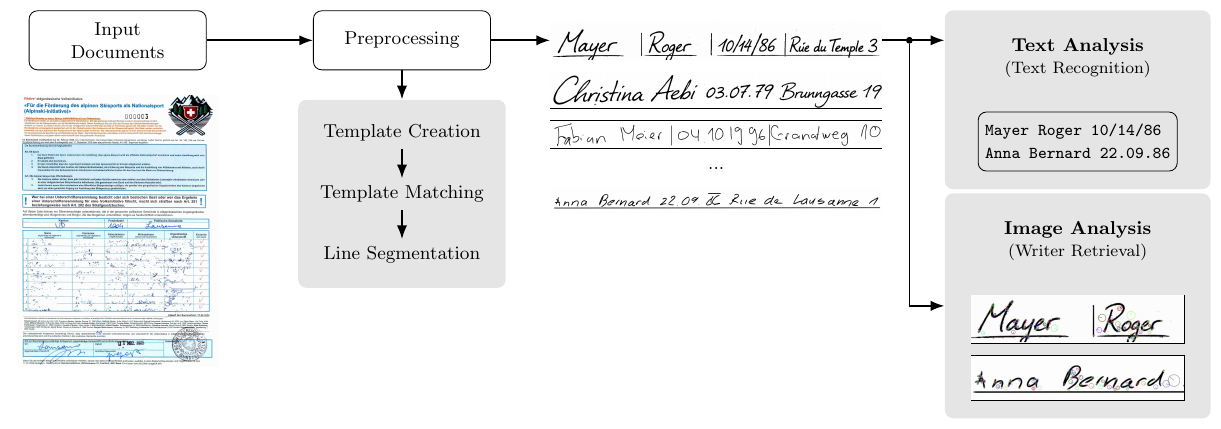}
    \caption{Overview of the methodology proposed. The handwriting on the form is blurred to preserve privacy.}
    \label{fig:placeholder}
\end{figure*}

\section{Introduction}


Popular initiatives constitute a central instrument of political participation in Switzerland, enabling citizens to transform societal concerns into legislative proposals. To submit a federal initiative, a predefined number of valid signatures must be collected from eligible voters. Because these signatures are typically recorded manually on paper forms, the verification of the collected data is an essential step in the validation process. Currently, this verification is largely performed manually by authorities, making it a time-consuming and labor-intensive procedure.

Given the number of submitted initiatives and the large volume of handwritten entries that must be processed, there is growing interest in technological solutions that can assist or partially automate this verification workflow. We address the detection of potential duplicates in two ways: text-based through OCR and image-based through writer retrieval. Recent advances in AI and OCR \cite{li2021trocr} provide opportunities to improve both the efficiency and reliability of the validation process. Secondly, methods for handwriting analysis and writer identification have been shown to identify similar handwriting styles in large collections of documents \cite{peer-diss,christlein_handwriting_2018}. Despite the potential of these approaches, the use of automated document analysis techniques for the validation of initiative signature lists has received little systematic investigation. 

The objective of our study is therefore to investigate whether and to what extent technical tools can support the automated analysis of initiative signature lists. In particular, the work focuses on methods for extracting handwritten entries from scanned documents and for identifying potential irregularities through text- and image-based handwriting comparison. The study addresses the following research questions:

\begin{itemize}
    \item Which methodological approaches are suitable for the automated extraction of handwritten entries from signature lists, and what challenges arise in this process?
\item How can AI-based methods be applied to identify duplicate entries and similar handwriting styles within signature lists, and how effective are these approaches?
\end{itemize}

To answer these questions, we analyze scanned signature lists originating from a simulated initiative that consists of 443 handwritten text lines and 418 writers. Our methodology proposed is shown in Fig. \ref{fig:placeholder}, including the extraction of the handwriting by using a template creation and matching algorithm, followed by traditional as well as more advanced AI-based models designed for text recognition and writer retrieval used for detecting similar handwriting. We find that the current state of the art of text recognition particularly struggle with text that is out-of-vocabulary - our best model achieves a CER of 29.6\,\% for recognition of first names. For writer retrieval, we achieve a mAP of 50.6\,\% and a Top-10 accuracy of 86.6\,\%, which shows that our system is able to find similar handwriting entries in signature lists, even when only one line of text is available.

The remainder of this paper is structured as follows. \Cref{sec:related_work} describes related work, while \cref{sec:dataset} introduces the dataset and  \cref{sec:methodology} presents the methodology, starting with the preprocessing, \ie template matching and line segmentation applied to the scanned documents. Afterwards, we discuss the OCR systems and the writer retrieval methods  used for finding duplicate entries on the text-based and image-based similarity. Finally, \cref{sec:results} discusses the results and \cref{sec:conclusion} summarizes the findings and outlines potential directions for future work.

\section{Related Work} \label{sec:related_work}

In this section, we briefly highlight the current research directions of the parts of our methodology.

\paragraph{Document Layout Analysis and Form Processing} Transformer-based models such as LayoutLM and its successors~\cite{layoutlm} have demonstrated strong performance in document understanding by jointly modeling text and layout information. Similarly, OCR-free approaches such as Donut~\cite{donut} directly parse documents into structured representations. Despite these advances, real-world applications involve highly structured forms with consistent layouts. In such cases, template-based approaches remain effective: Compared to generic document understanding models, template-based methods can exploit structural regularities and require less training data. Therefore, the problem setting of signature list analysis differs from general document understanding, as it combines structured layout processing with unconstrained handwritten content.

\paragraph{Handwritten Text Recognition}

Handwritten text recognition has transitioned from recurrent neural network-based approaches with CTC~\cite{ctc} to transformer-based architectures, such as TrOCR~\cite{li2021trocr}. Garrido et al.~\cite{GarridoMunoz2026} highlight that transformer models achieve state-of-the-art performance on standard benchmarks, but their effectiveness strongly depends on large-scale training data and implicit language modeling. In our practical scenario, however, the system must operate under limited data conditions and encounter out-of-vocabulary tokens such as names and addresses. As we show, this significantly degrades performance. More recently, open- and closed-source vision-language models have been explored for text recognition tasks, e.g., Qwen~\cite{qwen3_technical_report}, and show competitive performance (2\% CER on IAM)~\cite{llm-ocr}.

\paragraph{Writer Retrieval} Traditional approaches rely on handcrafted features such as SIFT descriptors combined with aggregation techniques like VLAD~\cite{christlein15ICDAR,peer_netmvlad}, which have proven effective for capturing local handwriting characteristics. 

Recent approaches employ neural networks with metric learning to learn discriminative representations of handwriting~\cite{keglevic,peer_netmvlad}. For historical data, self-supervised, \eg masked autoencoders or DINO-based training protocols~\cite{peer_saghog_2024,Raven2024}, as well as clustering-based~\cite{unsupervised_icdar17} approaches have emerged as popular alternatives.

Furthermore, handwriting comparison has long been studied in the context of forensic document examination. Systems such as CEDAR-FOX~\cite{Srihari2018} provide tools for assisting forensic experts in comparing handwriting samples. These systems typically rely on handcrafted features and expert-driven analysis and are designed for manual inspection rather than large-scale automated processing.

\section{Dataset} \label{sec:dataset}

\begin{figure}[h]
    \centering
        \subfloat{\includegraphics[width=0.48\linewidth]{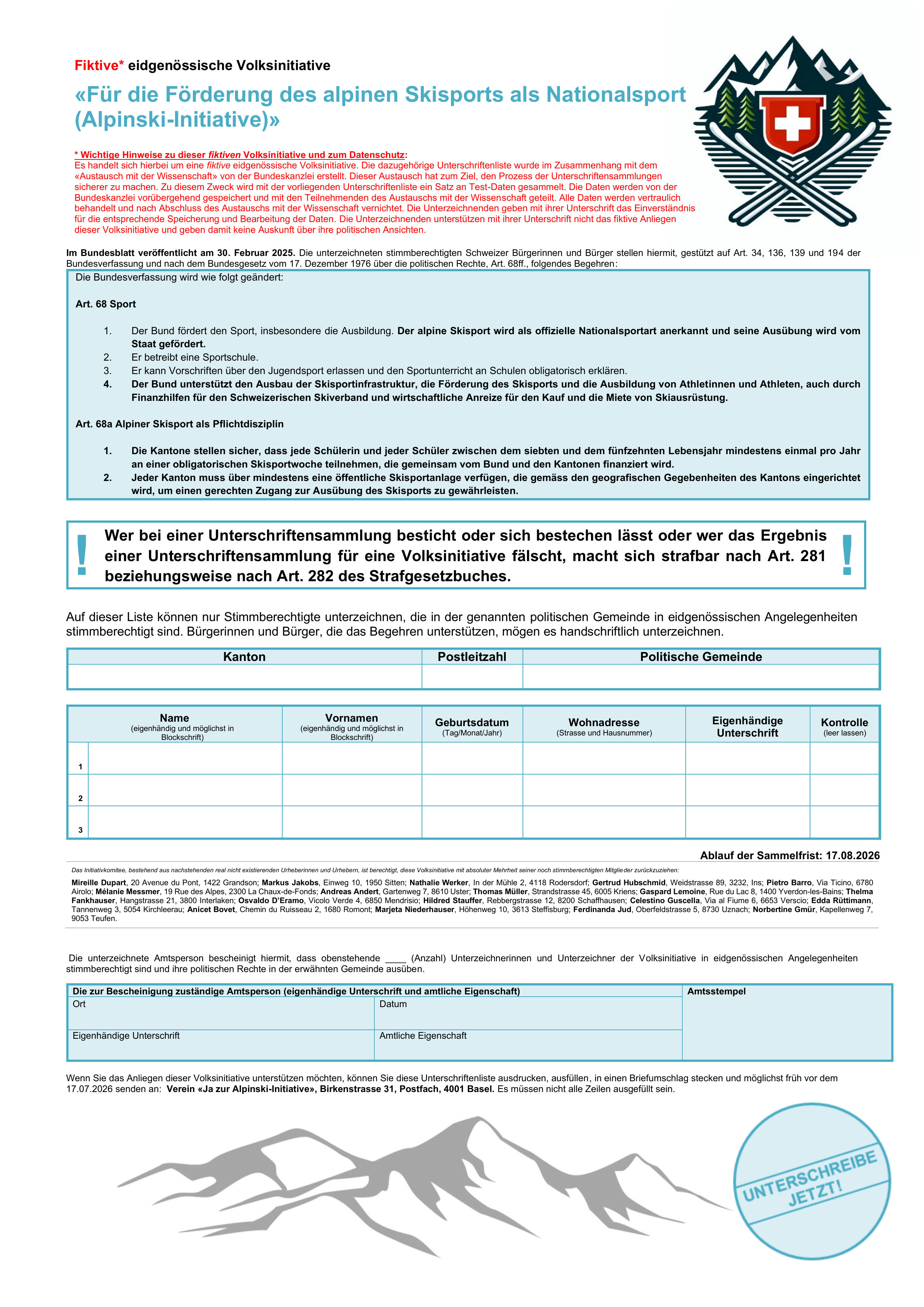}}
            \subfloat{\includegraphics[width=0.48\linewidth]{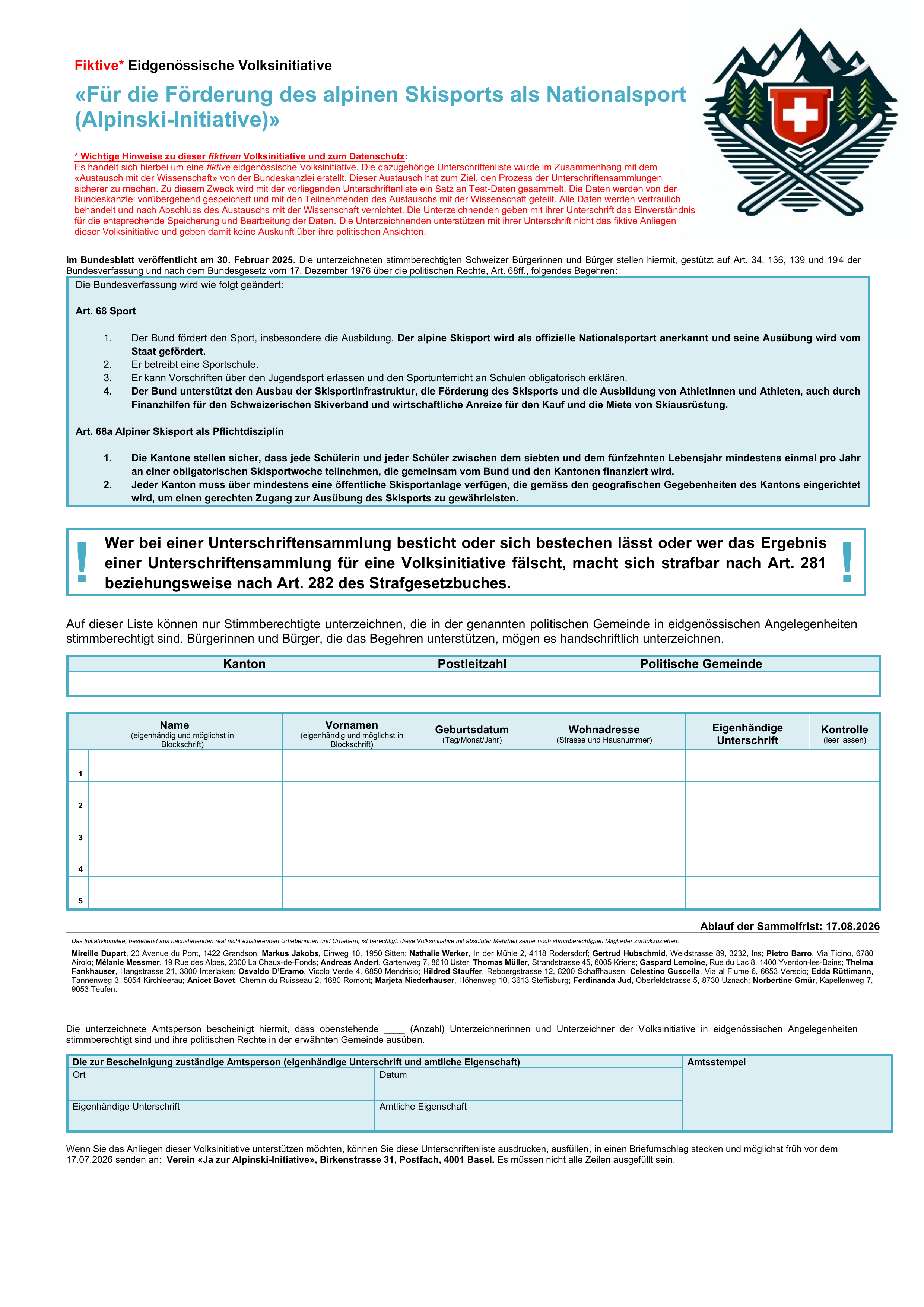}}

        \caption{Examples of the templates.}
        \label{fig:TestIniativeTemplates}
\end{figure}

We evaluate our method on a simulated Swiss popular initiative ``For the promotion of alpine skiing as a national sport (Alpine Ski Initiative)'', for which 418 writers contributed 443 text lines, denoted as \emph{Test-SL}. The initiative (with two templates shown in \cref{fig:TestIniativeTemplates}) follows a real one regarding the form and its layout and had already been evaluated by authorities to include annotations indicating the validity of each entry as well as the type of anomaly, if present. The statistics are shown in \cref{tab:writer_samples}. In addition, the printed and digitized documents contain visual artifacts such as rotations, scaling variations, and printing errors. Each entry typically contains the fields \emph{first name (Vorname)}, \emph{last name (Nachname)}, \emph{address}, \emph{date of birth}, and a signature, which constitutes as a single line of handwritten text.

Furthermore, we have received real signature lists that consist of about 100,000 text lines which we use for training the writer retrieval approaches (denoted as \emph{Real-SL}).

\begin{table}[h]
        \centering
        \captionof{table}{Writer statistics of the extracted lines}
        \label{tab:writer_samples}
        \begin{tabular}{lc}
            \toprule
            \textbf{Writer-ID} & Num. samples \\
            \midrule
            $\{1,2,3,4\}$ & 5 \\
            5             & 10 \\
            6--418 ("Singletons")        & 1 \\
            \midrule
            \textbf{Total} & \textbf{443} \\
            \bottomrule
        \end{tabular}
        \end{table}

\section{Methodology} \label{sec:methodology}
\subsection{Preprocessing}

Preprocessing consists of three steps: (i) template identification, (ii) template extraction, and (iii) line segmentation to obtain the handwritten entries of each form.

\subsubsection*{Template Identification}

Layout-based features are extracted using Histogram of Oriented Gradients (HOG)~\cite{dalal2005hog}. Before feature extraction, scans are grouped by orientation and nearly empty or corrupted pages are removed. These features are embedded with UMAP~\cite{mcinnes2018umap} and clustered using HDBSCAN~\cite{mcinnes2017hdbscan}, which handles an unknown number of templates and identifies outliers as noise.

To reduce over-segmentation, cluster centroids are compared using cosine similarity, and highly similar clusters are merged.

\subsubsection*{Template Extraction}

For each cluster, a high-quality reference scan is selected and all pages are aligned in a two-stage process: coarse alignment using SIFT features with RANSAC-based affine transformation estimation~\cite{lowe2004sift,fischler1981ransac}, followed by fine alignment via Enhanced Correlation Coefficient (ECC) optimization~\cite{Evangelidis08}.

A median image of the aligned scans is computed to suppress handwriting while preserving consistent structures. A stability mask identifies regions that remain constant across scans. The median template is used for a second alignment pass, alignment quality is assessed, and low-quality scans are removed.

Residual artifacts are detected using Median Absolute Deviation (MAD) and removed via inpainting. The final template is constructed from stable regions, with morphological operations compensating for minor misalignments.

\subsubsection*{Line Segmentation}

Accurate line segmentation requires knowledge of the table structure. An automated approach using the vision-language model Qwen3-VL-8B-Instruct~\cite{qwen3_technical_report} proved unreliable for column detection. Therefore, cell boundaries were annotated manually.

Each scan is first assigned to the best-matching template, and low-confidence assignments are excluded. Handwritten content is extracted by template subtraction. Each scan is aligned to its assigned template using the previously described two-stage registration process. The printed layout is removed using a template-derived mask applied to a binarized image, and noise is reduced through connected-component analysis and morphological filtering.

Residual printed lines are detected using Canny edge detection~\cite{canny} and the probabilistic Hough transform~\cite{hough1962method}, and removed only when they do not overlap with handwriting. Rows with insufficient ink are discarded. Final line images are obtained by masking the grayscale scan and normalizing the background, while metadata is inherited from the template annotations.

\subsection{Text Recognition}

Automatic text recognition on signature lists poses several challenges due to the characteristics of the data. In contrast to common handwriting recognition benchmarks, which typically consist of continuous text written under controlled conditions (e.g., IAM~\cite{marti2002iam}, CVL~\cite{cvl}), the examined documents primarily contain short sequences such as names, birth dates, and addresses. Moreover, many entries are written in uncontrolled environments, for example while standing in public spaces, which often results in irregular or difficult-to-read handwriting. This mismatch between typical training data and the target domain can negatively affect recognition performance. To assess the robustness of modern text recognition approaches under these conditions, we evaluate a set of OCR and handwriting recognition models that differ in architecture, language modeling capability, and level of recentness. The evaluated methods are briefly described in the following subsections.

\paragraph{Tesseract}

Tesseract~\cite{smith2009tesseract} is a widely used open-source OCR engine originally developed by Hewlett-Packard and later maintained by Google. It provides pretrained language models for a large number of languages and training periods. In this work, the German language model (\texttt{deu}) was used for transcription.

\paragraph{Kraken}

Kraken~\cite{kiessling2019kraken} is an OCR system designed for the recognition of historical and non-Latin scripts. It is based on neural sequence models and has been widely applied in historical document analysis. The selected model was trained on 20th-century French handwriting.

\paragraph{PaddleOCR}

PaddleOCR~\cite{du2020ppocr} is part of the PaddlePaddle deep learning ecosystem and provides a modular pipeline for document analysis. Although the framework supports multiple document understanding tasks, only the text recognition component was used in this study.

\paragraph{TrOCR}

TrOCR~\cite{li2021trocr} is a transformer-based OCR model developed by Microsoft. It demonstrates strong performance in handwritten text recognition tasks. However, due to its integrated language modeling capabilities, the system may occasionally produce linguistically plausible but contextually incorrect outputs when the visual evidence is uncertain. 

\paragraph{Qwen}

Qwen~\cite{qwen3_technical_report} is a recent family of multilingual large language models developed by Alibaba Cloud that can be executed locally. In this work, both Qwen2 and Qwen3 were evaluated across different model sizes. In addition, several prompting strategies were explored to generate text transcriptions.

\subsection{Writer Retrieval}

\begin{figure}[t!]
    \centering
    \includegraphics[width=0.75\linewidth]{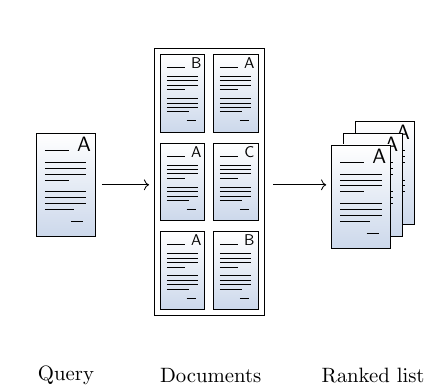}
    \caption{Writer Retrieval. Documents written by writer A should be found within the corpus using the query document.}
    \label{fig:wr_approach}
\end{figure}

The handwriting similarity is realized using algorithms for writer identification (also referred to as \emph{Writer Retrieval}). Fig.~\ref{fig:wr_approach} illustrates the basic principle of this task: each sample is used once as a so-called query, and the remaining entities in the dataset are ranked according to a similarity metric. The goal of writer retrieval is to compute an appropriate page descriptor (a vector) that represents the individual handwriting style and enables a reliable comparison between different writing samples.

Our methodology follows previous work presented in~\cite{peer2026}. Instead of relying on a keypoint detector, typically SIFT~\cite{unsupervised_icdar17,peer_netmvlad}, we sample keypoints directly along the contour of the binarized images (black pixels, as they represent the ink), since this improves performance, particularly when only little handwriting is present~\cite{christlein_handwriting_2018}. Using these keypoints, we compute SIFT descriptors and additionally extract $32\times32$ patches, which are subsequently fed into the neural network. Empty patches—those that do not contain handwriting—are discarded.
\paragraph{Local Feature Extraction}
We use two feature extractors: SIFT and a deep-learning variant. For SIFT descriptors, we use the RootSIFT descriptor, a modified version of the standard SIFT descriptor proposed by Arandjelovic et al.~\cite{arandjelovic_all_2013} to improve feature matching. It consists of an $l_1$ normalization, a square-root transformation—each element of the $l_1$-normalized vector is transformed by taking the square root—followed by an $l_2$ normalization. The descriptor angles are set to zero, making the descriptor variant with respect to rotations.

As a deep-learning-based feature extractor, ResNet56~\cite{he_deep_2016}, as used in related work~\cite{peer_netrvlad}, is employed. The network is trained on the $32\times32$ patches sampled along the contour of the handwriting. The linear layer is removed, and the embeddings are trained directly using triplet loss on semi-hard triplets. Two training targets are used:
\begin{itemize}
\item Supervised: The writer label is used as the target to train the network. Additionally, we employ instance supervision by assuming that each writer in the dataset provided by the Federal Chancellery contributed only a single sample; we refer to this as \emph{I-Sup}.
\item Cluster as a surrogate label (Cl-S): The SIFT descriptors of the sampled patches are clustered, and their cluster assignments are used as pseudo-labels to train the network~\cite{unsupervised_icdar17} in an unsupervised manner. The number of clusters is set to 5000.
\end{itemize}
\paragraph{Global Descriptor Formation} Writer retrieval is largely dominated by VLAD-based encoding methods, including both the classical formulation and its learnable extension, NetVLAD~\cite{netvlad}. In the original VLAD approach, a visual vocabulary is constructed by clustering (via $k$-means) local descriptors into a fixed number of centers. Each descriptor is assigned to its nearest center, and the residuals between descriptors and their assigned centers are accumulated. A variation, denoted as mVLAD, repeats this process (clustering and assignment) multiple times and concatenates the final vectors. Due to the employed hard assignment, the encoding is not differentiable.
NetVLAD replaces the hard assignment with a soft, differentiable alternative. Specifically, it computes assignment weights for each descriptor across all clusters using a learnable linear transformation followed by a softmax normalization. The residuals are then aggregated using these soft weights. Both the assignment parameters and the cluster centers are learned jointly, enabling end-to-end training within deep architectures.

\paragraph{Postprocessing} We aggregate all sampled embeddings of the handwritten line using $l_2$ normalization followed by sum pooling to obtain the global descriptor. Furthermore, we apply power normalization $f(x) = \text{sign}(x)\sqrt{|x|}$ followed by $l_2$ normalization. Finally, dimensionality reduction with whitening is performed using PCA. For ranking, the entities are ordered by computing the cosine similarity of the global descriptors.
\section{Results}\label{sec:results}

\paragraph{Text Recognition}

\begin{table}[t]
    \centering
    \caption{CER for the different entities. Lower values indicate better performance. The best results are highlighted in bold.}
    \label{tab:cer}
\begin{tabular}{lccccc}
    \toprule
        Method & Surname & Forename & Date & Address & Total \\
    \midrule
        TrOCR & 30.8 & 35.6 & 36.6 & 44.1 & 84.5 \\
        Tesseract & 79.3 & 78.6 & 66.2 & 80.3 & 95.5 \\
        Kraken & 76.0 & 71.3 & 46.8 & 71.1 & 91.1 \\
        PaddleOCR & 72.1 & 68.0 & 51.3 & 72.1 & 84.9 \\
        Qwen2-VL-7B & 57.2 & 64.3 & 24.0 & 65.7 & \textbf{76.9} \\
        Qwen3-VL-2B & 47.9 & 37.5 & 16.1 & 58.9 & 84.4 \\
        Qwen3-VL-8B & \textbf{29.6} & \textbf{23.3} & \textbf{6.3} & \textbf{33.6} & 80.4 \\
        Qwen3-VL-32B & 49.4 & 36.6 & 21.3 & 53.4 & 85.3 \\
    \bottomrule
\end{tabular}
\end{table}

The evaluated methods are assessed using the CER metric and are shown in Table \ref{tab:cer}.
Overall, the results show relatively high error rates compared to academic handwriting datasets. For example, TrOCR \cite{li2021trocr} reports approximately 3\,\% CER on the IAM dataset, while the evaluated dataset yields significantly higher error rates. This suggests that the dataset presents a more challenging recognition scenario, likely not only due to variations in handwriting, but also to the presence of out-of-vocabulary words like names or addresses. In contrast to standard benchmarks, where models can benefit from implicit or explicit language modeling, the evaluated setting requires recognition without relying on familiar word patterns, which we think substantially increases the difficulty of the task.

Furthermore, among the tested models, Qwen3-VL-8B achieved the best overall performance, particularly for the date (6.3\,\%), first name (23.3\,\%), and address (33.7\,\%) fields. It also obtained the lowest last-name CER (29.6\,\%), slightly outperforming TrOCR (30.8\,\%), which performs competitively on individual fields but shows a very high error rate when recognizing the entire line, suggesting that field-level evaluation provides a more meaningful analysis.

Traditional OCR systems such as Tesseract, Kraken, and PaddleOCR perform significantly worse, with CER values mostly between 66\,\% and 80\,\% for names and addresses. This may be explained by the fact that these systems are primarily trained on printed text rather than handwritten data.

\paragraph{Writer Retrieval}

The writer retrieval is evaluated in a leave-one-out scheme: Every sample is once used as a query and the remaining samples in the test set are ranked according to the cosine similarity of the global descriptors. Our metrics are the mean Average Precision (mAP), which takes into account the full ranked list as well as the soft Top-$x$ criteria, considering if at least one correct sample is among the first $x$ entries. We show the results in Table \ref{tab:wr_results}.

\begin{table*}
\centering
\caption{Results of different writer retrieval methods and datasets showing that a supervised use of additional training data is beneficial, which especially improved the ResNet\,+\,mVLAD system.}\label{tab:wr_results}
\begin{tabular}{llccccc}
\toprule
~ & Training data & mAP  & Top-1 & Top-3 & Top-5 & Top-10 \\ \midrule
SIFT + mVLAD & Test-SL & 41.9 & 70.6 & \textbf{82.0} & \textbf{82.6} & 84.0 \\
SIFT + mVLAD & Real-SL & 36.5 & 64.0 & 76.6 & 80.0 & 82.6\\ 
SIFT + mVLAD & CVL & 27.9 & 41.3 & 60.0 & 70.6 & 77.3\\ \midrule
ResNet + mVLAD (Cl-S) & Test-SL & 39.6 & 50.0 & 73.3 & 76.6 & 83.3\\ 
ResNet + mVLAD (Cl-S) & Real-SL & 49.3 & 66.6 & 80.0 & 80.0 & 83.3\\ 
ResNet + mVLAD (I-Sup.) & Real-SL & \textbf{50.6} & \textbf{76.6} & 80.0 & 80.0 & \textbf{86.6}\\ 
ResNet + mVLAD (Sup.) & CVL & 14.3 & 23.3 & 43.3 & 60.0 & 66.6\\ 
ResNet + mVLAD (Cl-S.) & CVL & 33.5 & 56.6 & 60.0 & 66.6 & 73.3\\ \midrule
ResNet + NetVLAD (Cl-S) & Test-SL & 18.4 & 33.3 & 63.3 & 73.3 & 80.0\\ 
ResNet + NetVLAD (Cl-S) & Real-SL & 16.9 & 26.6 & 60.0 & 60.0 & 70.0\\ 
ResNet + NetVLAD (I-Sup.) & Real-SL &  23.9 & 50.0 & 60.0 & 66.6 & 80.0\\ 
ResNet + NetVLAD (Sup.) & CVL & 19.8 & 43.3 & 63.3 & 70.0 & 73.3\\
ResNet + NetVLAD (Cl-S.) & CVL & 13.8 & 30.0 & 60.0 & 66.6 & 76.6\\ \midrule
ResNet (Cl-S.) & Test-SL & 24.7 & 40.0 & 50.0 & 60.0 & 73.3 \\
ResNet (Cl-S.) & Real-SL & 27.0 & 46.6 & 56.6 & 73.3 & 76.6 \\
ResNet (I-Sup.) & Real-SL & 30.1 & 60.0 & 70.0 & 73.3 & 83.3 \\
ResNet (Sup.) & CVL & 31.5 & 66.6 & 73.3 & 76.6 & 80.0 \\
ResNet (Cl-S) & CVL & 33.1 & 53.3 & 73.3 & 80.0 & 80.0 \\
\bottomrule
\end{tabular} 
\end{table*}

Overall, mVLAD aggregation consistently outperforms NetVLAD and non-aggregated ResNet features across the training datasets, particularly in terms of mAP and Top-1 accuracy.
With unsupervised training on the Test-SL, the SIFT + mVLAD approach achieves the best Top-3 (82.0\%) and Top-5 (82.6\%) accuracy, showing that handcrafted features combined with VLAD-aggregation remains competitive.
However, we achieve the best result when training on the real data (Real SL) --- which is significantly larger with around 100k text lines and achieves the highest mAP (50.6\%) and Top-1 accuracy (76.6\%). This indicates that deep features benefit significantly from mVLAD aggregation, especially when additional (pseu\-do)\-supervision is available. The best-performing method is also trained directly on the unlabeled target corpus, suggesting that domain adaptation is important.

In contrast, NetVLAD-based approaches consistently perform worse, yielding significantly lower mAP values across all datasets. This may be explained by the fact that NetVLAD is trained only once with a single codebook, which may reduce robustness and bias the representation toward the training distribution. We also observe that training on a publicly available dataset such as CVL \cite{cvl} leads to worse performance.

\begin{figure}
\centering
\input{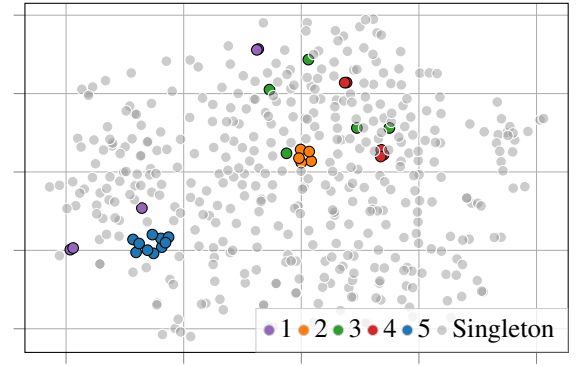}
\caption{t-SNE visualization of the writers showing that most writers's samples can be found in distinctive clusters.}
\label{fig:t-sne}
\end{figure}

Finally, we provide a t-SNE projection of the best-performing method (ResNet + mVLAD with instance supervision) in Fig. \ref{fig:t-sne}. It shows that the model is able to cluster writing samples from several writers, particularly writers 2, 4, and 5. However, some writers (e.g., writer 1 and writer 3) are not clearly separated in the feature space.

\section{Conclusion} \label{sec:conclusion}
This work demonstrates that the proposed pipeline is suitable for the automated analysis of handwritten signature lists. Handwritten content can be reliably segmented from scanned forms, and AI-based methods support the identification of duplicate entries and visually similar handwriting styles. The template-based subtraction achieves high precision for text extraction. While generic OCR systems exhibit higher error rates on short handwritten fields (e.g., names and addresses) due to out-of-vocabulary instances, performance can likely be improved through task-specific fine-tuning. In addition, publicly available address databases could be used for postprocessing to validate extracted entries.

For writer retrieval, similar entries are often ranked highly, providing a useful basis for further forensic inspection. However, robustness is limited by the limited amount of handwriting per entry and the sensitivity of similarity measures to identical or similar text content. 

Future work will focus on integrating writer retrieval and text recognition into a unified framework and on collecting a large-scale dataset to enable reproducible evaluation and further research.

\textbf{Acknowledgements} This research was undertaken as part of a project commissioned by the Swiss Federal Chancellery. The authors would like to express their gratitude for the provision of data and for the constructive feedback on the final manuscript.

\bibliographystyle{IEEEtran}
\bibliography{bib}

\end{document}